\documentclass[english, 10pt, twocolumn, twoside]{IEEEtran}
\usepackage{threeparttable}
\usepackage{mathrsfs}

\usepackage{cite}

\ifCLASSINFOpdf
\else
\fi
\usepackage{graphics}

\usepackage{subfigure}
\usepackage[cmex10]{amsmath}
\usepackage{amsfonts}
\usepackage{cases}
\usepackage{epsfig}
\usepackage{url}

\usepackage{algorithm}
\usepackage{algorithmic}

\usepackage{array}
\usepackage{amsmath}

\usepackage{multirow}
\usepackage{bigstrut}

\usepackage{bm}
\usepackage{amsmath}
\usepackage{amssymb}
\usepackage{mathrsfs}
\usepackage{color}

\usepackage{booktabs}
\usepackage{multirow}
\usepackage{amsmath}
\usepackage[bottom]{footmisc}
\usepackage{bbm}

\begin{document}

%
\title{StyleQoRA: Quality-Aware Low-Rank Adaptation for Few-Shot Multi-Style Editing}
%
%
%

\author{Cong Cao, Huanjing Yue, \textit{Senior Member, IEEE}, Yujie Xu, Xiaodong Xu
\thanks{C. Cao, Y. Xu are with the Imvision, SenseTime Group. X. Xu is with the ACE Robotics, SenseTime Group. H. Yue (corresponding author) is with the School of Electrical and Information Engineering, Tianjin University, Tianjin, China, 300072.}}

%
%

\markboth{}%
{Cao \MakeLowercase{\textit{et al.}}: StyleQoRA: Quality-Aware Low-Rank Adaptation for Few-Shot Multi-Style Editing}
%



\maketitle

\begin{abstract}
In recent years, image editing has garnered growing attention. However, general image editing models often fail to produce satisfactory results when confronted with new styles. The challenge lies in how to effectively fine-tune general image editing models to new styles using only a limited amount of paired data and a minimum number of parameters. To address this issue, this paper proposes a novel few-shot multi-style editing framework. For this task, we construct a benchmark dataset that encompasses five distinct styles. Correspondingly, we propose Quality-Aware Low-Rank Adaptation for few-shot multi-style editing (StyleQoRA). Our StyleQoRA can automatically determine the optimal rank for each layer through a novel approach that estimates the importance score of each single-rank component using an image quality metric. To balance specialization and knowledge sharing, we design a Mixture-of-Experts (MoE) LoRA with hybrid routing in our StyleQoRA, consisting of style-specific routing to prevent cross-style confusion and style-shared routing to capture common transformation patterns. Additionally, we explore the optimal location to insert LoRA within the Diffusion in Transformer (DiT) model and integrate adversarial learning and flow matching to guide the diffusion training process. Experimental results demonstrate that our proposed method outperforms existing state-of-the-art approaches with significantly fewer LoRA parameters.
\end{abstract}


%
\IEEEpeerreviewmaketitle

\section{Introduction}
\label{sec:intro}

With the rapid advancement of diffusion models, image editing has achieved remarkable progress. Existing general-purpose image editing methods \cite{brooks2023instructpix2pix,zhang2023magicbrush,hui2024hq,zhao2024ultraedit,shi2024seededit,huang2025photodoodle} are capable of performing diverse semantic modifications. However, style editing fundamentally differs from ordinary image editing. Unlike localized content manipulation, style editing requires learning consistent global and local transformation patterns—such as tone mapping, color/contrast/brightness adjustment, texture modulation, and stylization—while strictly preserving the original content. This makes style editing a challenging problem.

When encountering a previously unseen style, general editing models often fail to produce satisfactory results because the style distribution was not covered during training. Fine-tuning thus becomes necessary. However, paired editing data for a new style are typically scarce, making few-shot style editing both practically important and technically challenging. The difficulty lies not only in adapting to new styles with limited supervision but also in doing so in a parameter-efficient manner to avoid overfitting and redundancy.

\begin{figure}[H]
    \centering
    \includegraphics[width=1.0\linewidth]{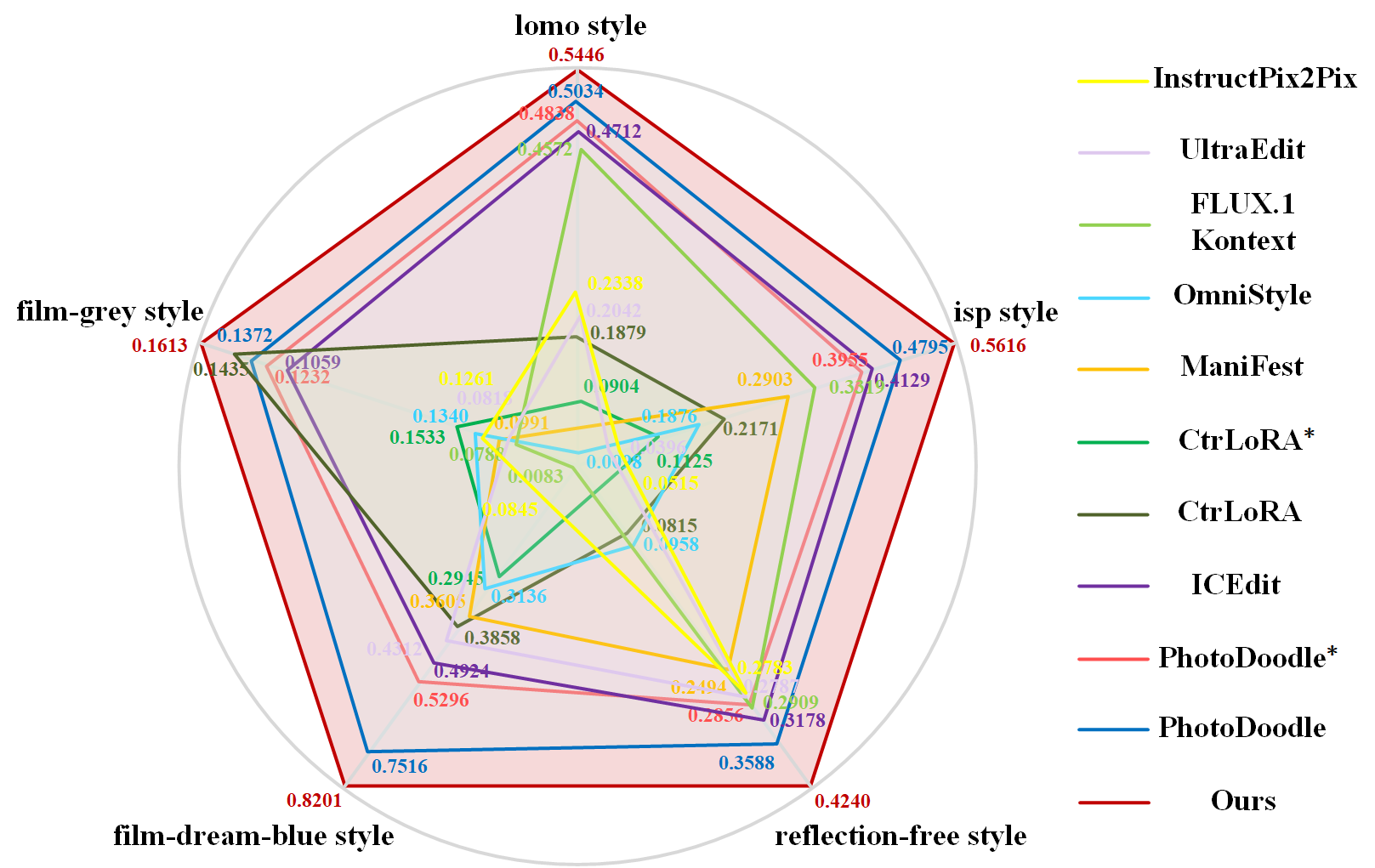}
    \caption{Comparison of style editing performance in terms of style similarity (SS) metric.}
    \label{fig:dataset}
\end{figure}

Low-Rank Adaptation (LoRA) provides a promising solution for parameter-efficient fine-tuning. However, directly applying fixed-rank LoRA to few-shot style editing is suboptimal. Different styles exhibit varying levels of complexity, and different layers of diffusion transformers contribute unevenly to style modeling. Therefore, using a fixed and manually selected rank inevitably leads to either underfitting or redundant parameters. We argue that the central challenge in few-shot multi-style editing is how to adaptively select and allocate LoRA ranks in order to achieve maximum performance gain with the minimum number of parameters.

To address this challenge, we propose StyleQoRA, a quality-aware low-rank adaptation framework tailored for few-shot multi-style image editing. Instead of assigning fixed ranks, we decompose high-rank LoRA layers into single-rank components and dynamically evaluate their importance during fine-tuning. Unlike prior work that measures importance using the Frobenius norm \cite{mao2024dora}, we observe that such norm-based criteria do not faithfully reflect the contribution of components in editing tasks. We therefore introduce a quality-driven evaluation strategy: by removing each single-rank component and measuring the degradation in image quality (e.g., PSNR) on a assessment image, we obtain a more accurate importance estimate. This enables adaptive rank allocation across layers and styles.

\begin{figure*}
    \centering
    \includegraphics[width=0.9\linewidth]{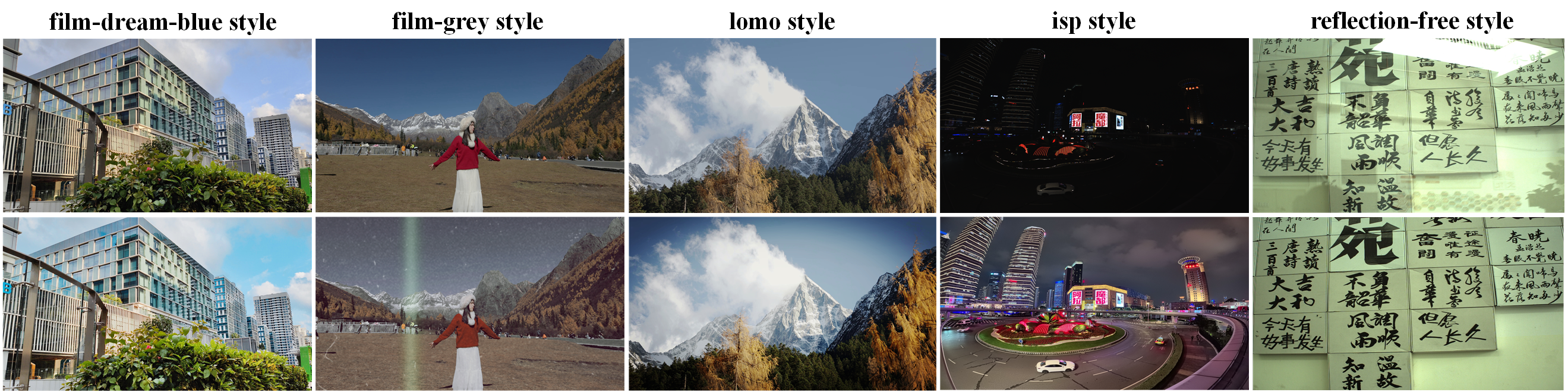}
    \caption{Examples of five styles in our dataset.}
    \label{fig:dataset}
\end{figure*}

We find that different styles have common patterns that can be learned together, and multi-style joint training can benefit from more training data compared with each single-style training. But multi-style joint training can introduce interference between styles. To balance specialization and knowledge sharing, we design a Mixture-of-Experts (MoE) LoRA with hybrid routing in our StyleQoRA. It consists of style-specific routing to prevent cross-style confusion and style-shared routing to capture common transformation patterns. This design further improves parameter efficiency and generalization.

In addition, since few-shot supervision makes style representation learning more challenging, we incorporate adversarial training into rectified flow to enhance stylistic fidelity. We also systematically investigate the optimal placement of LoRA modules in diffusion transformers and find that applying LoRA to single-stream transformer blocks achieves the best trade-off between performance and efficiency.

In a nutshell, our contributions can be summarized as follows:

\begin{itemize}
\item We define a new challenge task: few-shot multi-style image editing. To support this task, we constructed a benchmark dataset containing five representative styles. Accordingly, we propose a novel framework for this task.

\item We propose a quality-aware low-rank adaptation for efficient multi-style editing (StyleQoRA). Our framework decomposes high-rank LoRA layers into single-rank components and dynamically estimates their importance through image-quality-based evaluation instead of weight magnitude. This enables principled rank pruning and adaptive rank allocation across layers. To mitigate style interference during joint training, we further introduce a hybrid routing mechanism that combines style-specific routing for specialization and style-shared routing for capturing common transformation patterns.
 
\item We demonstrate that optimal rank allocation and routing design significantly improve parameter efficiency and performance. Our method systematically explores the optimal insertion locations for LoRA in diffusion transformers and integrates adversarial learning to enhance stylistic fidelity in few-shot settings. Extensive experiments show that our method outperforms existing state-of-the-art approaches while using only 3.7$\%$ of the LoRA parameters compared to PhotoDoodle.
\end{itemize}

\section{Related Work}

\subsection{Image Editing}

Recently, the advancement of the diffusion model has spurred the development of image editing \cite{brooks2023instructpix2pix,zhang2023magicbrush,hui2024hq,zhao2024ultraedit,shi2024seededit,batifol2025flux,chen2025unireal,yang2023eliminating,wang2025vimaedit,dai2025erddci,huang2025e4c,sun2026vision,huang2025photodoodle}. \cite{brooks2023instructpix2pix,zhang2023magicbrush,hui2024hq,zhao2024ultraedit,shi2024seededit} drive the development of general image editing by building larger and better image editing datasets. UniReal \cite{chen2025unireal} treats the input and output images in the image editing task as frames and learns general image editing from large-scale videos. FLUX.1 Kontext \cite{batifol2025flux} proposes a flow matching model that unifies image generation and editing by incorporating semantic context from text and image inputs. However, when these general image editing models encounter a specific new style that is not included in their training data, since the instructions cannot accurately describe the new style, these models cannot generate satisfactory results. The only way is to fine-tune these models on the data of the new style. Although PhotoDoodle \cite{huang2025photodoodle} proposes to fine-tune general image editing for photo doodling, it only focuses on the photo doodling style that only changes the local areas and neglects the global operations such as contrast, brightness, and tone styles. Our work is the first framework that focuses on more general few-shot image style editing, covering both global and local changes. 

\subsection{Few-Shot Image Generation}

Numerous customization methods for few-shot text-to-image generation already exist, such as Dreambooth \cite{ruiz2023dreambooth}, CustomDiffusion \cite{kumari2023multi}, and StyleDrop \cite{sohn2023styledrop}. Despite these achievements, a substantial gap still persists between few-shot text-to-image generation and few-shot image editing. Text-to-image generation solely focuses on the consistency between the generated image and the given prompt. However, image editing requires a balance between the consistency of the generated image with the prompt and the preservation of content. In the early stage, ManiFest \cite{pizzati2022manifest} proposed a framework for few-shot image translation. This framework utilizes adversarial learning to learn a context-aware representation of the target domain from a few images. CtrLora \cite{xu2025ctrlora} trains different LoRAs on a base ControlNet for few-shot controllable image generation. But CtrLora \cite{xu2025ctrlora} still requires hundreds of paired data for fine-tuning a new style. Our work only requires 41 pairs for fine-tuning. PhotoDoodle \cite{huang2025photodoodle} applies a plain LoRA to a pretrained denoising transformer for few-shot photo doodling but does not explore the efficiency of LoRA. In our work, we propose StyleQoRA which requires significantly fewer parameters than PhotoDoodle.

\subsection{Image Stylization}


In the initial stage, StyleClip \cite{patashnik2021styleclip} and StyleGAN-NADA \cite{gal2022stylegan} have demonstrated how text descriptors can adapt the style of source images via StyleGAN \cite{karras2019style,karras2020analyzing}. However, they can only be applied to several categories such as faces, animals, cars, and churches, which are supported by StyleGAN. In recent years, with the success of diffusion in various fields, diffusion-based image stylization has attracted more and more attention. Customization methods \cite{ruiz2023dreambooth, sohn2023styledrop} can customize the text-to-image diffusion model to generate images with specific styles through fine-tuning. However, they are not designed for style editing and cannot preserve the content of an input image. Style transfer methods \cite{wang2024instantstyle,wang2025omnistyle} can transfer the style from a single style image to the content image. Nevertheless, a single image cannot accurately describe a kind of style. The approaches closest to ours are CtrLora \cite{xu2025ctrlora} and PhotoDoodle, which can edit the input image to specific styles through fine-tuning LoRAs on a few paired style editing data. But our method requires less fine-tuning data and far fewer LoRA parameters. 

\begin{figure*}
    \centering
    \includegraphics[width=1.0\linewidth]{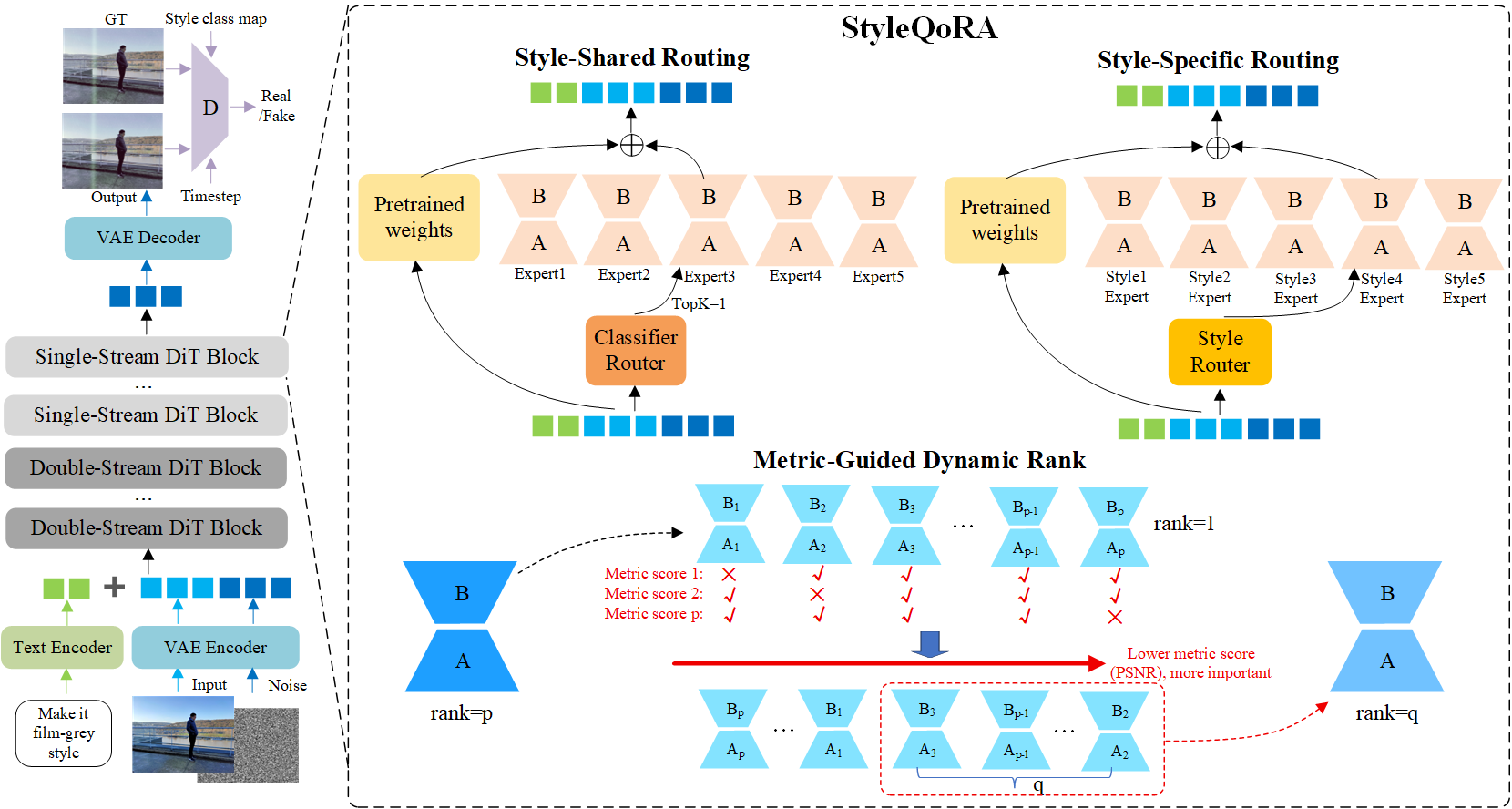}
    \caption{The framework of the proposed method. We propose StyleQoRA with style-specific and style-shared routing. Our StyleQoRA can automatically determine the optimal ranks for each layer with metric-guided dynamic rank.}
    \label{fig:framework}
\end{figure*}

\section{Dataset}

We construct a benchmark dataset with five different styles for few-shot multi-style editing. First, we collect 70 images from DSLR cameras, smartphones, and the Internet as the input for three styles (film-dream-blue, film-grey, and lomo styles). We use the software Meitu to generate the ground-truth images for these three styles. Then we collect 70 paired images from \cite{yao2025polarfree} to construct reflection-free style data. Additionally, we construct 70 paired images from a commercial ISP to create the ISP style. More specifically, we extract the input image after demosaicking the HDR raw data and obtain the ground-truth image from the final output of the ISP. Our styles have both global (color, contrast, and brightness) changes and local (texture) changes. For each style, 70 images are divided into a training set (41 images) and a testing set (29 images). Fig. \ref{fig:dataset} shows examples of five styles in our dataset. 


\section{Background}

\subsection{Flow Matching Model}

Given two data distributions $p_0$ and $p_1$ ($p_0$ denotes the target data distribution, and $p_1$ is the standard normal distribution $\mathcal{N}(0, 1)$), there exists a vector field $u_t$ that generates a probabilistic path $p_t$, which transitions from $p_0$ to $p_1$.

Following \cite{esser2024scaling}, we define the forward process as:
\begin{equation}
    x_t = a_t x_0 + b_t \epsilon, \quad \text{where } \epsilon \sim \mathcal{N}(0, 1)
\end{equation}
The coefficients $a_t$ and $b_t$ satisfy $a_0 = 1$, $b_0 = 0$, $a_1 = 0$, and $b_1 = 1$. They define a probabilistic path $p_t$ from $p_0$ to $p_1$. The transformed variable can be given by:
\begin{equation}
    x_t' = u_t(x_t | \epsilon) = \frac{a_t'}{a_t}x_t - \epsilon b_t(\frac{a_t'}{a_t}-\frac{b_t'}{b_t})
    \label{eq:coondition v}
\end{equation}
Flow matching trains a vector field $v_\theta(x, t)$, parameterized by a deep neural network, to approximate the marginal vector field $u_t(x_t|\epsilon)$. Therefore, flow matching minimizes the following objective \cite{lipman2022flow}:
\begin{equation}
\mathcal{L}_\text{CFM}(\theta) = \mathbb{E}_{t,\, p_t(x_t|\epsilon),\, p(\epsilon)} \left\| v_\theta(x_t, t) - u_t(x_t|\epsilon) \right\|^2
\end{equation}

\section{Method}

Given an input image $I_{in}$, we aim to map $I_{in}$ to the ground truth $I_{gt}$ with specific styles. Fig. \ref{fig:framework} presents the framework of the proposed method.

\subsection{StyleQoRA}

\subsubsection{Mixed Routing MoE LoRA}

Inspired by recent Mixture-of-Expert (MoE) works \cite{zhang2024clip,qu2024llama,zhu2024dynamic}, we propose an parameter-efficient multi-style MoE LoRA with a mixture of style-specific and style-shared routing to jointly fine-tune multiple styles.
The standard LoRA layer can be defined as:
\begin{equation}
\label{eq:lora_formula}
W = W_0 + \Delta W = W_0 + BA
\end{equation}

where $W_0$ and $W$ denote the weight matrices before and after fine-tuning, respectively. $A$ and $B$ denote the low-rank matrices. We combine LoRA with the MoE framework to enable expert LoRAs to adaptively learn which aspects to focus on, which can boost the model capacity without compromising computational efficiency. Each MoE LoRA layer contains $E$ LoRAs $\{W_{1}, \dots, W_{E}\}$ and a router that assigns the input $\mathbf{x}$ to experts according to style-shared or style-specific routing.

For style-shared routing, different styles are adaptively assigned shared LoRAs by a classifier router to learn common patterns. 
We utilize a classifier $W_z \in \mathbb{R}^{m \times E}$ to learn style-shared routing, the routing score for each expert can be defined as:
\begin{equation}
    p_{shared}^i(\mathbf{x}) = \frac{\exp(z^i(\mathbf{x}))}{\sum_{j=1}^{E} \exp(z^j(\mathbf{x}))}
\end{equation}
where $z(\mathbf{x}) = W_z \mathbf{x}$, $p^i(\mathbf{x})$ is the score for expert $i$. Let $\Omega_k(\mathbf{x})$ denote the indices of the top-$k$ scores, ensuring $|\Omega_k(\mathbf{x})| = k$ and $z^i(\mathbf{x}) > z^j(\mathbf{x})$ for all $i \in \Omega_k(\mathbf{x})$ and $j \notin \Omega_k(\mathbf{x})$. For style-shared routing, the weights for experts can be defined as:
\begin{equation}
    w_{shared}^i(\mathbf{x}) = 
    \begin{cases} 
        \frac{\exp(z^i(\mathbf{x}))}{\sum_{j \in \Omega_k(\mathbf{x})} \exp(z^j(\mathbf{x}))}, & \text{if } i \in \Omega_k(\mathbf{x}) \\
        0, & \text{otherwise}
    \end{cases}
    \label{eq:router}
\end{equation}
But we find that only style-shared routing will cause different styles to confuse with each other. Therefore, we propose style-specific routing to solve this problem. For style-specific routing, different styles are assigned independent LoRAs by the style router, which ensures that different styles do not confuse with each other. The routing score for each expert can be defined as: 
\begin{equation}
    p_{specific}^i(\mathbf{x}) = 
    \begin{cases} 
        1 & \text{if LoRA } W_i \text{ is assigned to i-th style} \\
        0, & \text{otherwise}
    \end{cases}
    \label{eq:router}
\end{equation}
Correspondingly, the weights for experts can be defined as:
\begin{equation}
    w_{specific}^i(\mathbf{x}) = 
    \begin{cases} 
        1 & \text{if LoRA }  W_i \text{ is assigned to i-th style} \\
        0, & \text{otherwise}
    \end{cases}
    \label{eq:router}
\end{equation}

We alternately assign style-shared routing and style-specific routing in a certain proportion. 
Each expert LoRA $W^i$ can be replaced by low-rank matrices $B^i$ and $A^i$:
\begin{align}
    \mathrm{MoE}_{\text{LoRA}}(\mathbf{x}) = W_0(\mathbf{x}) + \sum_{i=1}^E w^i(\mathbf{x}) \left( B^i A^i (\mathbf{x}) \right) \label{eq:moe_lora} 
\end{align}

\begin{table*}
\centering
\caption{Quantitative comparison with state-of-the-art methods for film-dream-blue style. The best results are highlighted in bold. }
\resizebox{0.7\textwidth}{21mm}{
\begin{tabular}{lccccccccc}
\toprule
Methods                          & PSNR$\uparrow$  & SSIM$\uparrow$  & $\Delta E_{ab}\downarrow$ & FID$\downarrow$  & LPIPS$\downarrow$ & CLIP-I$\uparrow$ & DINO $\uparrow$ &SS $\uparrow$ & CS $\uparrow$  \\
\hline
InstructPix2Pix &14.69 &0.5278 &32.57 &79.64  &0.4167 &0.9349 &0.8741 &0.0845 &0.4534 \\
UltraEdit       &13.29 &0.4229 &28.94 &213.42 &0.5580 &0.7545 &0.5661 &0.4312 &0.3309\\
FLUX.1 Kontext  &8.94  &0.2447 &60.85 &136.87 &0.5582 &0.8628 &0.7773 &0.0083 &0.2675\\
OmniStyle      &12.55 &0.4529 &30.74 &178.31 &0.5644 &0.8005 &0.6809 &0.3136 &0.3554\\
\hline
ManiFest        &14.63 &0.4461 &21.50 &191.99 &0.6844 &0.8022 &0.6423 &0.3605 &0.2922\\
CtrLoRA*        &16.80 &0.5820 &16.34 &62.07  &0.2182 &0.9499 &0.9386 &0.2945 &0.5560\\
CtrLoRA         &17.63 &0.5370 &14.51 &74.53  &0.2458 &0.9467 &0.9297 &0.3858 &0.5052\\
ICEdit          &19.62 &0.6222 &11.64 &49.10  &0.3412 &0.9641 &0.9586 &0.4924 &0.5367\\
PhotoDoodle*    &23.33 &0.8308 &7.43  &17.01  &0.0707 &0.9887 &0.9898 &0.5296 &0.8103\\
PhotoDoodle     &24.50 &0.8491 &6.92  &12.90  &0.0613 &0.9905 &0.9940 &0.7516 &0.8211\\
Ours            &\textbf{25.82} &\textbf{0.8580} &\textbf{5.78}  &\textbf{12.73}  &\textbf{0.0605} &\textbf{0.9917} &\textbf{0.9950} &\textbf{0.8201} &\textbf{0.8270}\\
\bottomrule
\end{tabular}
}
\label{ComparisonFilmdreamblue}
\end{table*}

\begin{table*}[t]
\centering
\caption{Quantitative comparison with state-of-the-art methods for film-grey style. The best results are highlighted in bold. }
\resizebox{0.7\textwidth}{21mm}{
\begin{tabular}{lccccccccc}
\toprule
Methods                          & PSNR$\uparrow$  & SSIM$\uparrow$  & $\Delta E_{ab}\downarrow$ & FID$\downarrow$  & LPIPS$\downarrow$ & CLIP-I$\uparrow$ & DINO $\uparrow$ &SS $\uparrow$ & CS $\uparrow$  \\
\hline
InstructPix2Pix &18.69 &0.5826 &16.50 &108.02 &0.3905 &0.9138 &0.8622 &0.1261 &0.4885 \\
UltraEdit       &13.46 &0.4160 &34.56 &264.02 &0.6385 &0.7513 &0.4621 &0.0818 &0.2960\\
FLUX.1 Kontext  &14.27 &0.3560 &22.94 &125.70 &0.4064 &0.8842 &0.8729 &0.0788 &0.2571\\
OmniStyle      &15.91 &0.5393 &18.80 &179.63 &0.5119 &0.8381 &0.7239 &0.1340 &0.4277\\
\hline
ManiFest        &17.83 &0.5729 &13.96 &119.19 &0.5796 &0.9118 &0.8341 &0.0991 &0.3871\\
CtrLoRA*        &17.31 &0.5567 &17.76 &116.20 &0.3324 &0.9108 &0.8690 &0.0533 &0.4944\\
CtrLoRA         &19.01 &0.5535 &11.96 &81.71  &0.2976 &0.9270 &0.9110 &0.1435 &0.4984\\
ICEdit          &20.57 &0.6325 &11.67 &74.22  &0.3710 &0.9287 &0.9144 &0.1059 &0.5239\\
PhotoDoodle*    &23.21 &0.8280 &7.59  &47.59  &0.1305 &0.9658 &0.9536 &0.1232 &0.7697\\
PhotoDoodle     &23.74 &0.8288 &6.77  &38.56  &0.1108 &\textbf{0.9759} &0.9738 &0.1372 &\textbf{0.7761}\\
Ours            &\textbf{24.14} &\textbf{0.8301} &\textbf{6.44}  &\textbf{36.41}  &\textbf{0.1095} &0.9742 &\textbf{0.9746} &\textbf{0.1613} &0.7751\\
\bottomrule
\end{tabular}
}
\label{ComparisonFilmgrey}
\end{table*}

\subsubsection{Metric-Guided Dynamic LoRA Rank}

High-rank LoRA can be decomposed into single-rank components:
\begin{equation}
W = W_0 + \sum_{k=1}^{r} \Delta W_k = W_0 + \sum_{k=1}^{r}B_kA_kc_k
\end{equation}
where $r$ denotes the rank of LoRA, i.e., the number of single-rank components. $A_k \in \mathbb{R}^{d\times1}$ and $B_k \in \mathbb{R}^{1\times d}$ are single-rank matrices. $c_k$ denotes a scalar, and it is set to 0 if the component is to be pruned. Then we can prune unimportant components according to the corresponding importance scores. 

Dora \cite{mao2024dora} utilizes the Frobenius norm to measure the importance score.
However, we find that the Frobenius norm cannot accurately measure the importance in the style editing task. In NLP tasks, the correlation between the total amount of parameter updates and task performance is relatively strong, making this criterion effective. However, in style editing, one component may occupy a significant portion of the parameter space yet primarily encode content perturbations or noise unrelated to style, while another component, though small in parameter magnitude, precisely carries key color or brushstroke signals. Therefore, we propose to use an image quality metric to estimate the importance score. First, we select one image from the training set as the LoRA assessment image. Let the complete denoising network be denoted as $N$. We remove each single-rank component $r_k$ and denote the residual network as $N_k$, and then predict the corresponding result. Finally, we calculate the PSNR metric between the result and the corresponding ground-truth image as the importance score. The lower the PSNR, the more important the removed component is. The proposed importance score can be formulated as:
\begin{equation}
\text{IS}(r_k) = M(N_k(I_{in}), I_{gt})
\label{eq:ab_lora}
\end{equation}
where $M$ denotes the metric function. To accelerate the calculation speed of the importance score, we only perform inference once from the noisy latents $x_t$ and directly estimate the clean latents $\hat{x}_0$ from $x_t$ as the result:
\begin{equation}
\hat{x}_0 = x_t - t v_\theta(x_t, t)
\label{eq:x0}
\end{equation}
Compared to the Frobenius norm, an important advantage of our method is its ability to identify redundancies: components whose removal leaves the metric nearly unchanged can be replaced by other components. Moreover, components deemed redundant on a single image are likely to remain redundant across multiple images, as redundancy arises from functional overlap within the parameter subspace rather than specific input conditions. Similarly, components identified as critical on a single image typically remain critical across multiple images. Using more LoRA assessment images would slightly improve performance, but since assessing all single-rank components is computationally expensive (e.g., taking several hours on an A100), we use only a single image for assessment. Throughout the training process, we first train our model for a period, then conduct an assessment, prune the less important single-rank components, and continue training.


To be specific, we divide all MoE LoRAs into different groups: a style-shared MoE LoRA group and several style-specific MoE LoRA groups, where each style constitutes a group.
For each single-rank component in the style-specific MoE LoRA of each style, we calculate the PSNR for the corresponding style as its importance score. For each single-rank component in the style-shared MoE LoRA, we calculate the sum of the PSNR values across all styles as its importance score. We then rank all importance scores within each group and prune the least important components using the same pruning ratio.

\subsubsection{LoRA Position Analysis}

We perform fine-tuning on our StyleQoRA using the pre-trained general image editing model of PhotoDoodle, which features a FLUX architecture. Inspired by \cite{frenkel2024implicit}, we explore the optimal position to insert LoRA into the FLUX architecture. Nevertheless, \cite{frenkel2024implicit} uses prompt injection to analyze the significance of LoRAs in different positions within SDXL \cite{podell2023sdxl}, a method that cannot be directly applied to the FLUX architecture. Specifically, prompt injection involves providing different prompts to LoRAs in various positions and measuring the importance of LoRA based on which one has a greater impact on the final outcome. However, prompt injection is based on the independence of text condition injection. In contrast, the injection of text conditions in FLUX does not show the same independence as in SDXL. The text conditions from the T5 encoder are injected into the first DiT block and generate new text conditions for the subsequent block. Therefore, we propose a new approach to analyze the importance of LoRAs in different positions. 

First, we insert LoRAs into various positions within the FLUX model and jointly fine-tune all LoRAs for few-shot style editing. Subsequently, we remove LoRAs from different positions of the fine-tuned model and predict the corresponding outcomes. Thereafter, we can analyze the significance of LoRA based on the impact of its removal. We find that, for the two types of blocks that constitute the FLUX model, namely the double-stream and single-stream denoising transformer blocks, LoRA applied to the double-stream denoising transformer block scarcely has any effect on the final result. We propose applying LoRA solely to the single-stream denoising transformer blocks, which can significantly reduce the number of LoRA parameters.

\begin{table*}[t]
\centering
\caption{Quantitative comparison with state-of-the-art methods for lomo style. The best results are highlighted in bold. }
\resizebox{0.7\textwidth}{21mm}{
\begin{tabular}{lccccccccc}
\toprule
Methods                          & PSNR$\uparrow$  & SSIM$\uparrow$  & $\Delta E_{ab}\downarrow$ & FID$\downarrow$  & LPIPS$\downarrow$ & CLIP-I$\uparrow$ & DINO $\uparrow$ &SS $\uparrow$ & CS $\uparrow$  \\
\hline
InstructPix2Pix  &16.15 &0.5325 &17.61 &75.84  &0.3328 &0.9456 &0.8995 &0.2338 &0.4658\\
UltraEdit        &14.59 &0.4823 &23.22 &173.00 &0.4782 &0.8456 &0.7269 &0.2042 &0.3880\\
FLUX.1 Kontext   &12.77 &0.3142 &22.23 &62.26  &0.3121 &0.9494 &0.9347 &0.4572 &0.2614\\
OmniStyle       &11.11 &0.3798 &30.95 &197.60 &0.6967 &0.7045 &0.6240 &0.0098 &0.3550 \\
\hline
ManiFest         &13.94 &0.4427 &23.07 &191.95 &0.6613 &0.7746 &0.6337 &0.2903 &0.2993 \\
CtrLoRA*         &18.30 &0.6167 &13.98 &53.04  &0.1789 &0.9617 &0.9555 &0.0904 &0.5978\\
CtrLoRA          &17.58 &0.5936 &13.78 &46.55  &0.1735 &0.9691 &0.9570 &0.1879 &0.5921\\
ICEdit           &20.35 &0.6467 &9.86  &42.06  &0.3112 &0.9694 &0.9617 &0.4712 &0.5593\\
PhotoDoodle*     &22.87 &0.8448 &8.34  &17.42  &0.0732 &0.9902 &0.9911 &0.4838 &0.8167\\
PhotoDoodle      &23.18 &0.8433 &7.29  &14.35  &\textbf{0.0598} &0.9917 &0.9920 &0.5034 &0.8170\\
Ours             &\textbf{23.73} &\textbf{0.8473} &\textbf{6.64}  &\textbf{13.77}  &0.0609 &\textbf{0.9924} &\textbf{0.9929} &\textbf{0.5446} &\textbf{0.8179}\\
\bottomrule
\end{tabular}
}
\label{ComparisonLomo}
\end{table*}

\begin{table*}[t]
\centering
\caption{Quantitative comparison with state-of-the-art methods for isp style. The best results are highlighted in bold. }
\resizebox{0.7\textwidth}{21mm}{
\begin{tabular}{lccccccccc}
\toprule
Methods                          & PSNR$\uparrow$  & SSIM$\uparrow$  & $\Delta E_{ab}\downarrow$ & FID$\downarrow$  & LPIPS$\downarrow$ & CLIP-I$\uparrow$ & DINO $\uparrow$ &SS $\uparrow$ & CS $\uparrow$  \\
\hline
InstructPix2Pix  &15.29 &0.4806 &21.83 &163.74 &0.5033 &0.8496 &0.7795 &0.0515 &0.4413 \\
UltraEdit        &13.07 &0.4056 &26.04 &274.44 &0.6856 &0.6680 &0.3995 &0.0396 &0.3250\\
FLUX.1 Kontext   &14.31 &0.3370 &20.93 &92.52  &0.3460 &0.9211 &0.9031 &0.3319 &0.3454\\
OmniStyle       &14.14 &0.4334 &25.00 &273.54 &0.6308 &0.7426 &0.5868 &0.1876 &0.3488\\
\hline
ManiFest         &14.90 &0.4244 &22.45 &229.05 &0.7050 &0.7813 &0.6262 &0.2494 &0.2687\\
CtrLoRA*         &17.32 &0.4769 &16.94 &106.26 &0.3278 &0.9134 &0.8913 &0.1125 &0.4805\\
CtrLoRA          &17.54 &0.5034 &16.27 &94.08  &0.2816 &0.9362 &0.9114 &0.2171 &0.5208\\
ICEdit           &20.28 &0.5973 &10.91 &72.09  &0.3907 &0.9530 &0.9325 &0.4129 &0.4818\\
PhotoDoodle*     &21.13 &0.6768 &11.08 &82.82  &0.3117 &0.9460 &0.9350 &0.3955 &0.5932\\
PhotoDoodle      &22.21 &0.7065 &9.46  &\textbf{55.01}  &0.2396 &0.9595 &\textbf{0.9629} &0.4795 &0.6433 \\
Ours             &\textbf{22.62} &\textbf{0.7286} &\textbf{9.25}  &57.00  &\textbf{0.2140} &\textbf{0.9663} &0.9621 &\textbf{0.5616} &\textbf{0.6611}\\
\bottomrule
\end{tabular}
}
\label{ComparisonISP}
\end{table*}

\subsection{Loss}

Besides the objective loss $\mathcal{L}_\text{CFM}$ in flow matching, we propose to introduce adversarial loss to better capture the patterns in different styles with limited data. We design a discriminator $D_\psi$. Besides the image input, $D_\psi$ also takes both the style class map $c$  and the timestep $t$ as conditions to better discriminate the results at different timesteps and for different styles. For the style condition, we extend the style class to the class map and concatenate the input image with the class map. For the timestep condition, we utilize a linear layer to predict the scaling and shift values to modulate the feature. For the adversarial loss, we utilize the Relativistic GAN framework \cite{jolicoeur2018relativistic} and employ zero-centered gradient penalties in R3GAN \cite{huang2024gan} to stabilize GAN training for limited data. For $x^t$ during training, we predict $x_0$ from $x^t$ by Eq. \ref{eq:x0}. Then we decode $x_0$ using the FLUX VAE decoder $\mathcal{D}$ and apply the adversarial loss. Given real data $x\sim p_\mathcal{R}$ and fake data $x\sim p_\theta$ generated by $G_\theta$, the discriminator loss is defined as: 
\begin{equation}
\label{eq:ganloss}
\mathcal{L}_{D}=\mathbb{E}\left[f\left(  D_\psi(I_{gt}, c, t)-D_\psi(\mathcal{D}(x_0), c, t) \right)\right]
\end{equation}
The function $f$ can be defined as:
\begin{equation}
f(x) = -log(1+e^{-x})
\end{equation}
The adversarial loss of the generator assumes a symmetrical form:
\begin{equation}
\label{eq:ganloss}
\mathcal{L}_{G}=\mathbb{E}\left[f\left(  D_\psi(\mathcal{D}(x_0), c, t)-D_\psi(I_{gt}, c, t) \right)\right]
\end{equation}
For zero-centered gradient penalties, we apply $R_1$ and $R_2$ when training the discriminator:
\begin{equation}
\begin{aligned}
R_1 &=\frac{\gamma}{2}\mathbb{E}_{x\sim p_\mathcal{R}}\left[\left\| \nabla_x D_\psi \right \|^2\right]\\ 
R_2 &=\frac{\gamma}{2}\mathbb{E}_{x\sim p_\theta}\hspace{0.06cm}\left[\left\| \nabla_x D_\psi \right \|^2\right]
\end{aligned}
\end{equation}
$R_1$ penalizes the gradient norm of $D_\psi$ on real data, and $R_2$ penalizes the gradient norm of $D_\psi$ on fake data, $\gamma$ is a hyper-parameter.

Besides adversarial loss, we also apply reconstruction loss $\mathcal{L}_{rec}$ and cosine color loss $\mathcal{L}_{color}$. These three losses serve as extra guided losses to guide the flow matching diffusion training.
\begin{equation}
\begin{split}
\label{eq:ganloss}
\mathcal{L}_{rec}=&\|(\mathcal{D}(x_0)-I_{gt}\|_1  \\
\mathcal{L}_{color}=&C(\mathcal{D}(x_0), I_{gt})
\end{split}
\end{equation}
where $C$ denotes cosine similarity.
The total loss for the generator can be formulated as:
\begin{equation}
\mathcal{L}_{total} = \mathcal{L}_\text{CFM} + \lambda_1 \mathcal{L}_\text{adv} + \lambda_2 \mathcal{L}_\text{rec} + \lambda_3 \mathcal{L}_\text{color}
\end{equation}
where $\lambda_1,\lambda_2,\lambda_3$ are hyper-parameters.

\section{Experiments}

\subsection{Training Details}

For isp style, since the HDR raw data passing through demosaicking are very dark, we convert it to a log image and directly feed it into the method to ensure that the accuracy in the dark area is not compromised. We utilize the pre-trained generative image editing method in PhotoDoodle \cite{huang2025photodoodle} as the backbone and fine-tune it with our method. For our StyleQoRA, the number of experts is set to 5, and TopK in style-shared routing is set to 1. The batch size is set to 1. The training iteration is set to 30000. The hyperparameters $\lambda_1,\lambda_2,\lambda_3$, $\gamma$ are set to 1, 1, 10, and 0.5, respectively. The proposed model is implemented in PyTorch and trained with an 80G NVIDIA A100 GPU. 

\begin{table*}[t]
\centering
\caption{Quantitative comparison with state-of-the-art methods for reflection-free style. The best results are highlighted in bold. }
\resizebox{0.7\textwidth}{21mm}{
\begin{tabular}{lccccccccc}
\toprule
Methods                          & PSNR$\uparrow$  & SSIM$\uparrow$  & $\Delta E_{ab}\downarrow$ & FID$\downarrow$  & LPIPS$\downarrow$ & CLIP-I$\uparrow$ & DINO $\uparrow$ &SS $\uparrow$ & CS $\uparrow$  \\
\hline
InstructPix2Pix  &13.63 &0.4322 &25.67 &356.93 &0.5925 &0.6749 &0.2955 &0.2783 &0.3234\\
UltraEdit        &14.54 &0.5659 &20.61 &257.78 &0.4592 &0.8055 &0.6195 &0.2787 &0.4392\\
FLUX.1 Kontext   &14.19 &0.3960 &17.52 &138.57 &0.3418 &0.9222 &0.8591 &0.2909 &0.2495 \\
OmniStyle       &13.58 &0.5303 &37.66 &268.03 &0.5332 &0.8063 &0.6357 &0.0958 &0.4557\\
\hline
ManiFest         &12.97 &0.4833 &26.13 &286.89 &0.5667 &0.8053 &0.5930 &0.3558 &0.3310\\
CtrLoRA*         &14.18 &0.5650 &26.14 &173.68 &0.3466 &0.8810 &0.8035 &0.0206 &0.5368\\
CtrLoRA          &16.31 &0.6075 &19.18 &145.26 &0.2653 &0.9060 &0.8563 &0.0815 &0.5992\\
ICEdit           &22.17 &0.7458 &\textbf{8.18}  &51.27  &0.2058 &0.9463 &0.9548 &0.3178 &0.7054 \\
PhotoDoodle*     &22.25 &0.8196 &8.53  &46.09  &0.1005 &0.9594 &0.9443 &0.2856 &0.8271\\
PhotoDoodle      &22.52 &0.8175 &8.27  &\textbf{36.09}  &0.0912 &0.9715 &0.9688 &0.3588 &0.8255\\
Ours             &\textbf{22.53} &\textbf{0.8207} &8.49  &41.79  &\textbf{0.0863} &\textbf{0.9735} &\textbf{0.9700} &\textbf{0.4240} &\textbf{0.8321}\\
\bottomrule
\end{tabular}
}
\label{ComparisonReflectionfree}
\end{table*}

\begin{table*}[t]
\centering
\caption{LoRA Parameters comparison with state-of-the-art methods. The best results are highlighted in bold. }
\resizebox{0.55\textwidth}{5.0mm}{
\begin{tabular}{lcccccc}
\toprule
Methods                          & CtrLoRA*  & CtrLoRA & ICEdit & PhotoDoodle*  & PhotoDoodle & Ours  \\
\hline
LoRA Params (M) $\downarrow$  & 1244.7 & 6223.5 &115.0 &358.6 &1793.1 &\textbf{66.7}\\
\bottomrule
\end{tabular}
}
\label{ComparisonParams}
\end{table*}

\begin{table*}
\centering
\caption{Ablation study for proposed LoRA Position Selection (LPS), Multi-style MoE LoRA (MML), Metric-guided Dynamic Rank (MDR), and Extra Guided Loss (EGL) in StyleQoRA. We also carried out an ablation study on Style-Shared Routing (SSHR) and Style-Specific Routing (SSPR) in MML. Moreover, we compared the Metric-guided Dynamic Rank (MDR) with the Norm-guided Dynamic Rank (NDR).}
\resizebox{0.525\textwidth}{27mm}{
\begin{tabular}{ccccccccc}
\toprule
\multicolumn{2}{l}{LPS}                  &$\times$     &$\checkmark$ &$\checkmark$ &$\checkmark$ &$\checkmark$ &$\checkmark$ &$\checkmark$\\\hline
\multirow{2}{*}{MML} &SSHR               &$\times$     &$\times$     &$\checkmark$ &$\checkmark$ &$\checkmark$ &$\checkmark$ &$\checkmark$\\\cline{2-9}
                     &SSPR               &$\times$     &$\times$     &$\times$     &$\checkmark$ &$\checkmark$ &$\checkmark$ &$\checkmark$\\\hline
\multicolumn{2}{l}{NDR}                  &$\times$     &$\times$     &$\times$     &$\times$     &$\checkmark$ &$\times$     &$\times$\\\hline
\multicolumn{2}{l}{MDR}                  &$\times$     &$\times$     &$\times$     &$\times$     &$\times$     &$\checkmark$ &$\checkmark$\\\hline
\multicolumn{2}{l}{EGL}                  &$\times$     &$\times$     &$\times$     &$\times$     &$\times$     &$\times$     &$\checkmark$\\\hline
\multicolumn{2}{l}{PSNR$\uparrow$}       &23.33        &23.37        &23.59        &24.98        &25.16        &25.68        &25.82            \\
\multicolumn{2}{l}{SSIM$\uparrow$}       &0.8308       &0.8294       &0.8439       &0.8533       &0.8532       &0.8571       &0.8580       \\
\multicolumn{2}{l}{$\Delta E_{ab}\downarrow$} &7.43    &7.39         &9.07         &6.19         &6.13         &5.92         &5.78         \\
\multicolumn{2}{l}{FID$\downarrow$}      &17.01        &18.78        &20.40        &12.88        &12.82        &12.75        &12.73              \\
\multicolumn{2}{l}{LPIPS$\downarrow$}    &0.0707       &0.0755       &0.0740       &0.0615       &0.0619       &0.0599       &0.0605              \\
\multicolumn{2}{l}{CLIP-I$\uparrow$}     &0.9887       &0.9834       &0.9846       &0.9912       &0.9905       &0.9916       &0.9917               \\
\multicolumn{2}{l}{DINO$\uparrow$}       &0.9898       &0.9900       &0.9909       &0.9944       &0.9940       &0.9948       &0.9950             \\
\multicolumn{2}{l}{SS$\uparrow$}         &0.5296       &0.5301       &0.5886       &0.7626       &0.7206       &0.8125       &0.8201           \\
\multicolumn{2}{l}{CS$\uparrow$}         &0.8103       &0.7979       &0.8025       &0.8249       &0.8258       &0.8266       &0.8270           \\
\multicolumn{2}{l}{LoRA Params (M)$\downarrow$}&358.6  &89.7         &89.3         &87.7         &66.7         &66.7         &66.7                  \\
\bottomrule
\end{tabular}
}
\label{Ablation}
\end{table*}

\subsection{Comparison with State-of-the-art Methods}

To demonstrate the effectiveness of the proposed method for few-shot image style editing, we compare it with state-of-the-art methods on our proposed dataset. The compared methods include general editing methods InstructPix2Pix \cite{brooks2023instructpix2pix}, UltraEdit \cite{zhao2024ultraedit}, FLUX.1 Kontext \cite{batifol2025flux}, ICEdit \cite{zhang2025context}, the style transfer method OmniStyle \cite{wang2025omnistyle}, the controllable image generation method CtrLoRA \cite{xu2025ctrlora}, the few-shot image translation method ManiFest \cite{pizzati2022manifest}, and the few-shot image editing method PhotoDoodle \cite{huang2025photodoodle}. Besides InstructPix2Pix, UltraEdit, FLUX.1 Kontext, and OmniStyle, all methods are fine-tuned on our dataset. For CtrLoRA and PhotoDoodle, they trained a LoRA for each style in their paper. For a fair comparison, we also train them jointly on all styles and utilize style prompts to distinguish different styles. The jointly trained CtrLoRA and PhotoDoodle are named CtrLoRA* and PhotoDoodle*, respectively. For InstructPix2Pix, UltraEdit, and FLUX.1 Kontext, we give the corresponding detailed style descriptions when testing each style. For the style transfer method OmniStyle, for each style, an image is randomly selected from the corresponding training ground truth as the style image, while the input images are used as the content images.  

\begin{figure*}[t]
    \centering
    \includegraphics[width=0.91\linewidth]{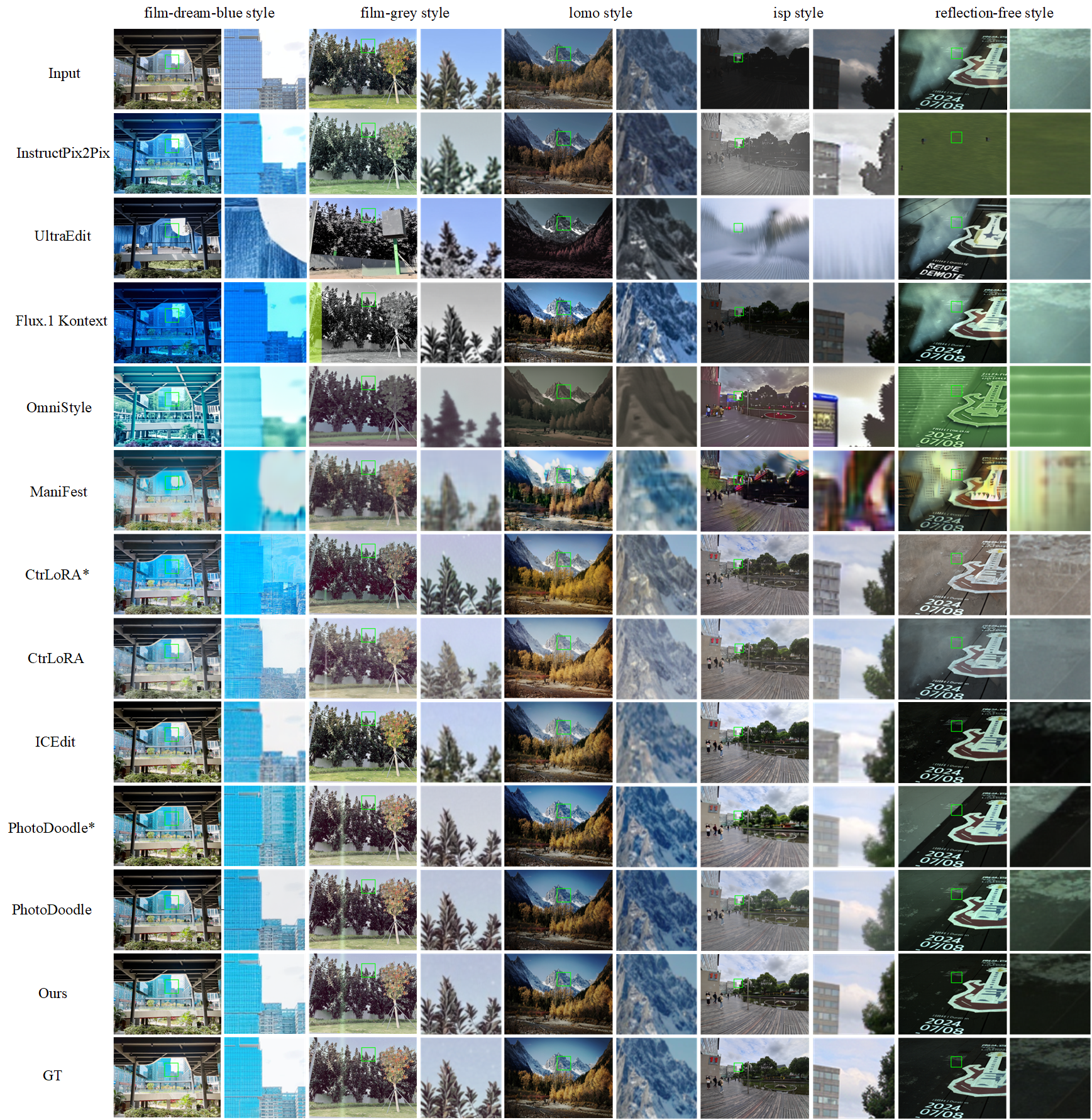}
    \caption{Visual quality comparison on our dataset. Zoom in for better observation}
    \label{fig:compare}
\end{figure*}

We utilize nine metrics to measure the image quality. These metrics include PSNR, SSIM, the $L_2$-distance in the CIE LAB color space ($\Delta E_{ab}$) \cite{zeng2020learning}, FID, LPIPS, CLIP image similarity (CLIP-I) \cite{zhao2024ultraedit}, DINO similarity (DINO) \cite{zhao2024ultraedit}, style similarity (SS) \cite{ke2023neural}, and content similarity (CS) \cite{ke2023neural}. For all reference-based metrics, we compute them between the results and the ground-truth images. 

Tables \ref{ComparisonFilmdreamblue}, \ref{ComparisonFilmgrey}, \ref{ComparisonLomo}, \ref{ComparisonISP}, \ref{ComparisonReflectionfree} list the style editing results for five styles. Table \ref{ComparisonParams} lists the total LoRA parameters of different methods. It can be observed that our method can outperform all state-of-the-art methods with fewer LoRA parameters. Take the film-dream-blue style as an example, our method outperforms the second best method, PhotoDoodle, in all nine metrics with only 3.7$\%$ of the LoRA parameters. Compared with PhotoDoodle, our method achieves 1.32 dB gain for PSNR, 0.0089 gain for SSIM, 1.14 gain for $\Delta E_{ab}$, 0.17 gain for FID, 0.0008 gain for LPIPS, 0.0012 gain for CLIP-I, 0.0010 gain for DINO similarity, 0.0685 gain for style similarity, and 0.0059 gain for content similarity. 

\begin{figure*}[t]
    \centering
    \includegraphics[width=0.91\linewidth]{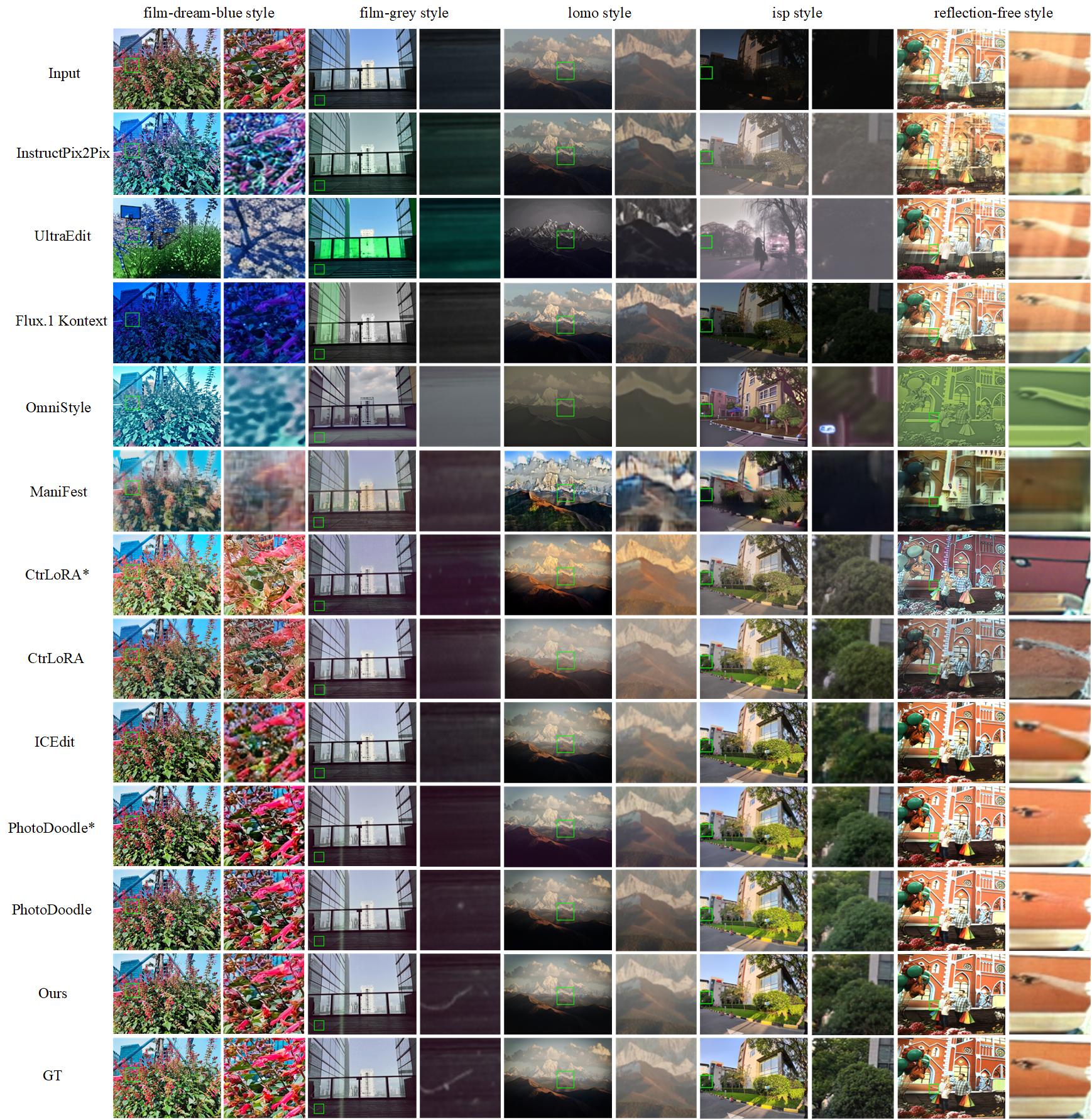}
    \caption{More visual quality comparison on our dataset. Zoom in for better observation}
    \label{fig:compare2}
\end{figure*}

Fig. \ref{fig:compare} presents the visual comparison results for five styles. Taking the film-dream-blue style as an example, it can be observed that our method is closest to the ground truth. PhotoDoodle, ICEdit, CtrLoRA, and ManiFest have color shifts. ICEdit, CtrLoRA, and ManiFest cannot preserve the content of the input image well. The results of PhotoDoodle* and CtrLoRA* are interfered with by another style (the reflection-free style) and generate wrong textures in highlight areas. Although detailed style descriptions are provided to InstructPix2Pix, UltraEdit, and FLUX.1 Kontext, text alone is insufficient to precisely characterize the style. Consequently, these general image editing techniques are unable to generate the desired style accurately. While a single image can convey the style more precisely than text, it still falls short of providing a completely accurate representation. The style transfer method, OmniStyle, focuses on the dominant color and texture of the style image as the defining characteristics of the style, yet this approach also fails to generate the appropriate style. For other styles, the result of our method is also the closest to the ground truth. Fig. \ref{fig:compare2} presents more comparison results. Taking the reflection-free style as an example, it can be observed that our method is closest to the ground truth. PhotoDoodle, PhotoDoodle*, CtrLoRA*, and ManiFest mistakenly identify the arm as a reflection and erase it. ICEdit and CtrLoRA cannot preserve the structure of the arms well. CtrLoRA, CtrLoRA*, ManiFest, and OmniStyle have severe color shifts. InstructPix2Pix, UltraEdit, and FLUX.1 Kontext cannot remove the reflection successfully. 

\subsection{Ablation Study}

In this section, we conduct an ablation study to demonstrate the effectiveness of the proposed LoRA Position Selection (LPS), Multi-style MoE LoRA (MML), Metric-guided Dynamic Rank (MDR), and Extra Guided Loss (EGL). For LoRA Position Selection, we only apply LoRA to single-stream denoising transformer blocks through LoRA Position Analysis. Regarding Multi-style MoE LoRA, we reduce the rank of each LoRA from 128 to 25. The number of experts is 5. We also carry out an ablation study on Style-Shared Routing (SSHR) and Style-Specific Routing (SSPR). For Metric-guided Dynamic Rank, we compare it with Norm-guided Dynamic Rank (NDR) which uses the Frobenius norm to measure the importance score \cite{mao2024dora}. For Extra Guided Loss (EGL), we apply adversarial loss, reconstruction loss, and cosine color loss to guide the diffusion training. Taking the film-dream-blue style as an example, Table \ref{Ablation} lists the quantitative comparison results by adding these modules one by one. 

It can be observed that when LoRA Position Selection is added, the LoRA parameters can be reduced to nearly 1/4, but the metrics can remain nearly the same. When Multi-style MoE LoRA with style-specific routing is added, the PSNR can be improved by 0.22 dB, and the style similarity can be improved by 0.0585. However, the interference between different styles makes the $\Delta E_{ab}$ and FID worse. By adding style-specific routing, the interference between different styles can be solved, and all metrics are significantly improved. The PSNR can be improved by 1.39 dB, the $\Delta E_{ab}$ can achieve a gain of 2.88, the FID can achieve a gain of 7.52, the LPIPS can achieve a gain of 0.0125, the CLIP-I can achieve a gain of 0.0066, and the style similarity can achieve a gain of 0.174. Although norm-guided dynamic rank can achieve improvement in a few metrics, it also makes the SSIM, LPIPS, CLIP-I, DINO similarity, and style similarity worse. Compared with norm-guided dynamic rank, metric-guided dynamic rank can better measure the importance of the single-rank component and achieve significantly better results in all metrics. When adding metric-guided dynamic rank, the PSNR can be improved by 0.7 dB, and the style similarity can achieve a gain of 0.0499. Moreover, dynamic rank can effectively further reduce the number of LoRA parameters. It is worth noting that we also attempted to directly reduce the rank without dynamic rank, which resulted in significantly worse performance compared to the original rank. Finally, Extra Guided Loss can further improve the performance, bringing 0.14dB gain for PSNR, 0.14 gain for $\Delta E_{ab}$, and 0.0076 gain for style similarity.

\section{Conclusion}

In this paper, we propose a novel framework for few-shot style editing. We propose StyleQoRA with style-specific and style-shared routing for jointly fine-tuning multiple styles. Our StyleQoRA can automatically estimate the importance score of each single-rank component using an image quality metric and determine the optimal ranks for each layer. Experimental results demonstrate that our method outperforms existing state-of-the-art methods with significantly fewer LoRA parameters, has only 3.7$\%$ of the LoRA parameters compared to PhotoDoodle.

%

%

\ifCLASSOPTIONcaptionsoff
  \newpage
\fi



\bibliographystyle{IEEEtran}
\bibliography{egbib}
\end{document}